\documentclass[11pt,a4paper]{article}
\usepackage[hyperref]{emnlp2020}
\usepackage{times}
\usepackage{latexsym}
\usepackage{tabularx}

\usepackage{bm}
\usepackage{times}
\usepackage{latexsym}
\usepackage{graphicx}
\usepackage{enumitem}
\usepackage{xcolor,colortbl}
\usepackage{tikz}
\usepackage{diagbox}
\usepackage{tkz-tab}
\usepackage{caption}
\usepackage{latexsym}
\usepackage{amssymb}
\usepackage{amsmath}
\usepackage{caption}
\usepackage{natbib}
\usepackage{blindtext}
\usepackage{url}
\usepackage{color}

\usepackage{url}
\usepackage{hyperref}    
\usepackage{amsthm}   
\usepackage{cleveref}

\usepackage{times}
\usepackage{latexsym}

\def\@fnsymbol#1{\ensuremath{\ifcase#1\or *\or \dagger\or \ddagger\or
   \mathsection\or \mathparagraph\or \|\or **\or \dagger\dagger
   \or \ddagger\ddagger \else\@ctrerr\fi}}

\usepackage{microtype}
\usepackage{amsmath}
\usepackage{float}
\usepackage[caption = false]{subfig}
\usepackage[T1]{fontenc} 
\usepackage[utf8]{inputenc}
\usepackage{booktabs} 
\usepackage{appendix,lipsum}

\usepackage{hyperref}

\usepackage{microtype}

\aclfinalcopy 
\def\aclpaperid{2284} 

\newcommand\blfootnote[1]{%
  \begingroup
  \renewcommand\thefootnote{}\footnote{#1}%
  \addtocounter{footnote}{-1}%
  \endgroup
}

\title{On Long-Tailed Phenomena in Neural Machine Translation}

\author{Vikas Raunak$^{1}$\qquad Siddharth Dalmia$^1$\qquad Vivek Gupta$^2$\qquad Florian Metze$^1$\\
$^1$~Carnegie Mellon University, USA \\
$^2$~University of Utah, USA \\
\texttt{\{vraunak,sdalmia,fmetze\}@cs.cmu.edu}\quad \texttt{vgupta@cs.utah.edu}}

\date{}

\begin{document}
\maketitle
\begin{abstract}
State-of-the-art Neural Machine Translation (NMT) models struggle with generating low-frequency tokens, tackling which remains a major challenge. The analysis of long-tailed phenomena in the context of structured prediction tasks is further hindered by the added complexities of search during inference. In this work, we quantitatively characterize such long-tailed phenomena at two levels of abstraction, namely, token classification and sequence generation. We propose a new loss function, the Anti-Focal loss, to better adapt model training to the structural dependencies of conditional text generation by incorporating the inductive biases of beam search in the training process. We show the efficacy of the proposed technique on a number of Machine Translation (MT) datasets, demonstrating that it leads to significant gains over cross-entropy across different language pairs, especially on the generation of low-frequency words. We have released the code to reproduce our results.\blfootnote{The first author is now a researcher at Microsoft, USA.}\footnote{\url{https://github.com/vyraun/long-tailed}} %
\end{abstract}

\section{Introduction}
\label{sec:intro}
Autoregressive sequence to sequence (seq2seq) models such as Transformers \cite{transformer} are trained to maximize the log-likelihood of the target sequence, conditioned on the input sequence. Furthermore, approximate inference (search) is typically done using the beam search algorithm \cite{reddy1988foundations}, which allows for a controlled exploration of the exponential search space. However, seq2seq models (or structured prediction models in general) suffer from a discrepancy between token level classification during \textit{learning} and sequence level inference during \textit{search}. This discrepancy also manifests itself in the form of the curse of sentence length i.e. the models' proclivity to generate shorter sentences during inference, which has received considerable attention in the literature \cite{curse, length_bias}.

In this work, we focus on how to better model long-tailed phenomena, i.e. predicting the long-tail of low-frequency words/tokens \cite{zhao2012long}, in seq2seq models, on the task of Neural Machine Translation (NMT). Essentially, there are two mechanisms by which tokens with low frequency receive lower probabilities during prediction: firstly, the norms of the embeddings of low frequency tokens are smaller, which means that during the dot-product based softmax operation to generate a probability distribution over the vocabulary, they receive less probability. This has been well known in Image Classification \cite{iclr2020} and Neural Language Models \cite{stolen}. Since NMT shares the same dot-product softmax operation, we observe that the same phenomenon holds true for NMT as well. For example, we observe a Spearman’s Rank Correlation of 0.43 between the norms of the token embeddings and their frequency, when a standard transformer model is trained on the IWSLT-14 De-En dataset
(more details in section \ref{sec:quantify}). Secondly, for transformer based NMT, the embeddings for low frequency tokens lie in a different subregion of space than semantically similar high frequency tokens, due to the different rates of updates \cite{frage_paper}, thereby, making rare words token embeddings ineffective. Since these token embeddings have to match to the context vector for getting next-token probabilities, the dot-product similarity score is lower for low frequency tokens, even when they are semantically similar to the high frequency tokens. 

Further, better modeling long-tailed phenomena has significant implications for several text generation tasks, as well as for compositional generalization \cite{compositionality_lake, raunak2019compositionality}. To this end, we primarily ask and seek answers to the following two fundamental questions in the context of NMT:\begin{enumerate}
\setlength{\itemsep}{0.25pt}
\item To what extent does better modeling long-tailed token classification improve inference?
\item How can we leverage intuitions from beam search to better model token classification?
\end{enumerate}
By exploring these questions, we arrive at the conclusion that the widely used cross-entropy (CE) loss limits NMT models' expressivity during inference and propose a new loss function to better incorporate the inductive biases of beam search.

\section{Characterizing the Long-Tail}
\label{sec:quantify}
In this section, we quantitatively characterize the long-tailed phenomena under study at two levels of abstraction, namely at the level of token classification and at the level of sequence generation. To illustrate the phenomena empirically, we use a six-layer Transformer model with embedding size 512, FFN layer dimension 1024 and 4 attention heads trained on the IWSLT 2014 De-En dataset \cite{cettolo2014report}, with cross-entropy and label smoothing of 0.1, which achieves a BLEU score of 35.14 on the \textit{validation set} using a beam size of 5.

\begin{figure}[ht!]
    \centering
    \vspace{-0.75em}
    \includegraphics[width=0.47\textwidth,height=0.38\textwidth]{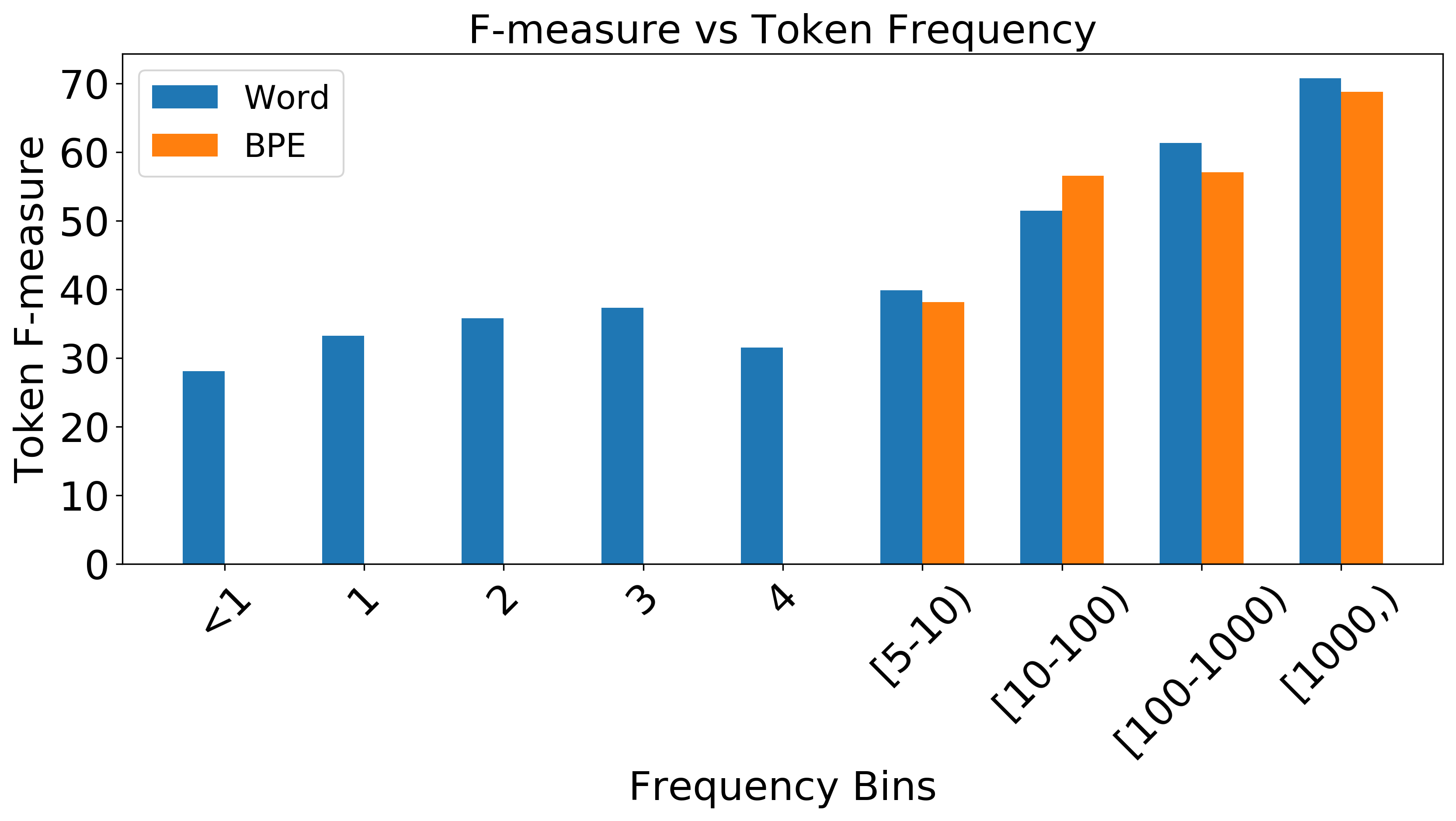} 
    \vspace{-0.5em}
    \caption{Token F-measure bucketed by Frequency: F-measure correlates with the tokens' training frequency.}
    \label{fig:figure1}
    \vspace{-1.5em}
\end{figure}
\subsection{Token Level}
At the token level, Zipf's law \cite{zipf} serves as the primary culprit for the long-tail in word distributions, and consequently, for sub-word (such as BPE \cite{bpe}) distributions . Figure \ref{fig:figure1} shows the F-measure \cite{neubig2019compare} of the target tokens bucketed by their frequency in the training corpus, as evaluated on the \textit{validation set}. Clearly, for tokens occurring only a few times, the F-measure is considerably lower for both words and subwords, demonstrating that the model isn't able to effectively generate low-frequency tokens in the output. Next, we study how this phenomenon is exhibited at the sequence (sentence) level.

\subsection{Sequence Level}
\label{seqlevel}
To quantify the long-tailed phenomena manifesting at the sentence level, we define a simple measure named the Frequency-Score, $F_{S}$ of a sentence, computed simply as the average frequency of the tokens in the sentence. Precisely, for a sequence $x$ comprising of $N$ tokens $[ x_{1}, \ldots, x_{i}, \ldots, x_{N} ]$, we define the Frequency-Score $F_{S}$ as: $F_{S}(x) = \cfrac{\sum_{i=1}^{N} f(x_{i})}{N}$, where $f(x_{i})$ is the frequency of the token $x_{i}$ in the training corpus. We compute $F_{S}$ for each source sequence in the IWSLT 2014 De-En \textit{validation set}, and split it into three parts of $2400$ sentences each, in terms of decreasing $F_{S}$ of the source sequences. The splits are constructed by dividing the validation set into three \textit{equal} parts based on the Frequency-score, so that we can compare the performance between the three splits for a given model.

Table \ref{tab:table1} shows the model performance on the three splits. Scores for 3 widely used MT metrics \cite{multeval}: BLEU, METEOR and TER  as well as the Recall BERT-Score (R-BERT) \cite{bert_score} are reported. The arrows represent the direction of better scores. The table shows that model performance across all metrics deteriorates as the mean $F_{S}$ value, $\hat{F_{S}}$ of the split decreases. On aggregate, this demonstrates that the model isn't able to effectively handle sentences with low $F_{S}$.

\begin{table}[t]
  \centering
  \small
  \setlength\tabcolsep{2.0pt}
  \begin{tabular}{l c c c c c}
    \toprule
    \textbf{Split} & \textbf{$ \cfrac{\boldsymbol{\hat{F_{S}}}}{10^4}  $} & \textbf{BLEU $\uparrow$} & \textbf{METEOR $\uparrow$} & \textbf{TER $\downarrow$} & \textbf{R-BERT $\uparrow$} \\
    \midrule
    Highest & 7.8 & 38.6 & 36.4 & 41.0 & 65.2 \\
    Medium & 5.3 & 34.1 & 34.2 & 45.5 & 61.0 \\
    Least & 3.4 & 33.0 & 34.1 & 46.2 & 60.6 \\
    \midrule
  \end{tabular}
  \vspace{-0.5em}
  \caption{Sequence Level Long-Tailed Phenomena: The performance across different metrics deteriorates with the mean Frequency-Score $\hat{F_{S}}$.}
  \label{tab:table1}
  \vspace{-1.5em}
\end{table}

\section{Related Work}
\label{sec:related}
At a high level, we categorize the solutions to better model long-tailed phenomena into three groups, namely, learning better representations, improving (long-tailed) classification and improvements in sequence inference algorithms. In this work, we will be mainly concerned with the \textit{interaction} between (long-tailed) classification and sequence inference.

\paragraph{Better Representations} Many recent works \cite{neubig, frage_paper, incorporating_bert} propose to either learn better representations for low-frequency tokens or to integrate pre-trained representations into NMT models. To better capture long range semantic structure, \citet{optimal_transport} argue for sequence level supervision during learning.

\paragraph{Long-Tailed Classification} A number of works, \cite{focal_loss, iclr2020}, have focused on designing algorithms that improve classification of low-frequency classes. Below, we list two such algorithms, used as baselines in section \ref{sec:section5}:

\paragraph{\textit{Focal Loss}} Proposed in \cite{focal_loss}, Focal loss (FL) increases the relative loss of low-confidence predictions vis-à-vis high confidence predictions, when compared to cross-entropy. It is described in equation \ref{focal_loss}, where $\gamma>0$ and $p$ refers to the probability/confidence of the prediction.
\begin{equation} \label{focal_loss}
\vspace{-0.25em}
\text{FL}(p) = -(1-p)^\gamma \log(p)
\end{equation}
\paragraph{\textit{$\tau$-Normalization}} \citet{iclr2020} link the norms of the penultimate (pre-softmax) layer to the frequency of the class in image classification (also shown to be true in the context of language models \cite{stolen}), and show that normalizing their weights $w_{i}$ i.e. leads to improved classification: 
\vspace{-1.0em}
\begin{equation}
\vspace{-0.25em}
\label{tau-norm}
\tilde{w_{i}} = \cfrac{w_{i}}{\lvert\lvert w_{i} \rvert\rvert^{\tau}} 
\end{equation}
Here, $\tau$ is a hyperparameter. The intuition behind $\tau$-Normalization is based on the simple observation that the norms of the penultimate layer dictate the feature span of the corresponding class during prediction.

At the sequence level, a parallel line of work has explored penalizing overconfident predictions \cite{entropy_regularization}, e.g., Label smoothing has been shown to yield consistent gains in seq2seq tasks \cite{label_smoothing_hinton}. 

\paragraph{Sequence Inference} \citet{diverse_beam_search, optimal_beam} try to modify beam search to allow for better exploring the output state space.

\begin{figure}[ht!]
       \includegraphics[width=0.48\textwidth,height=0.31\textwidth]{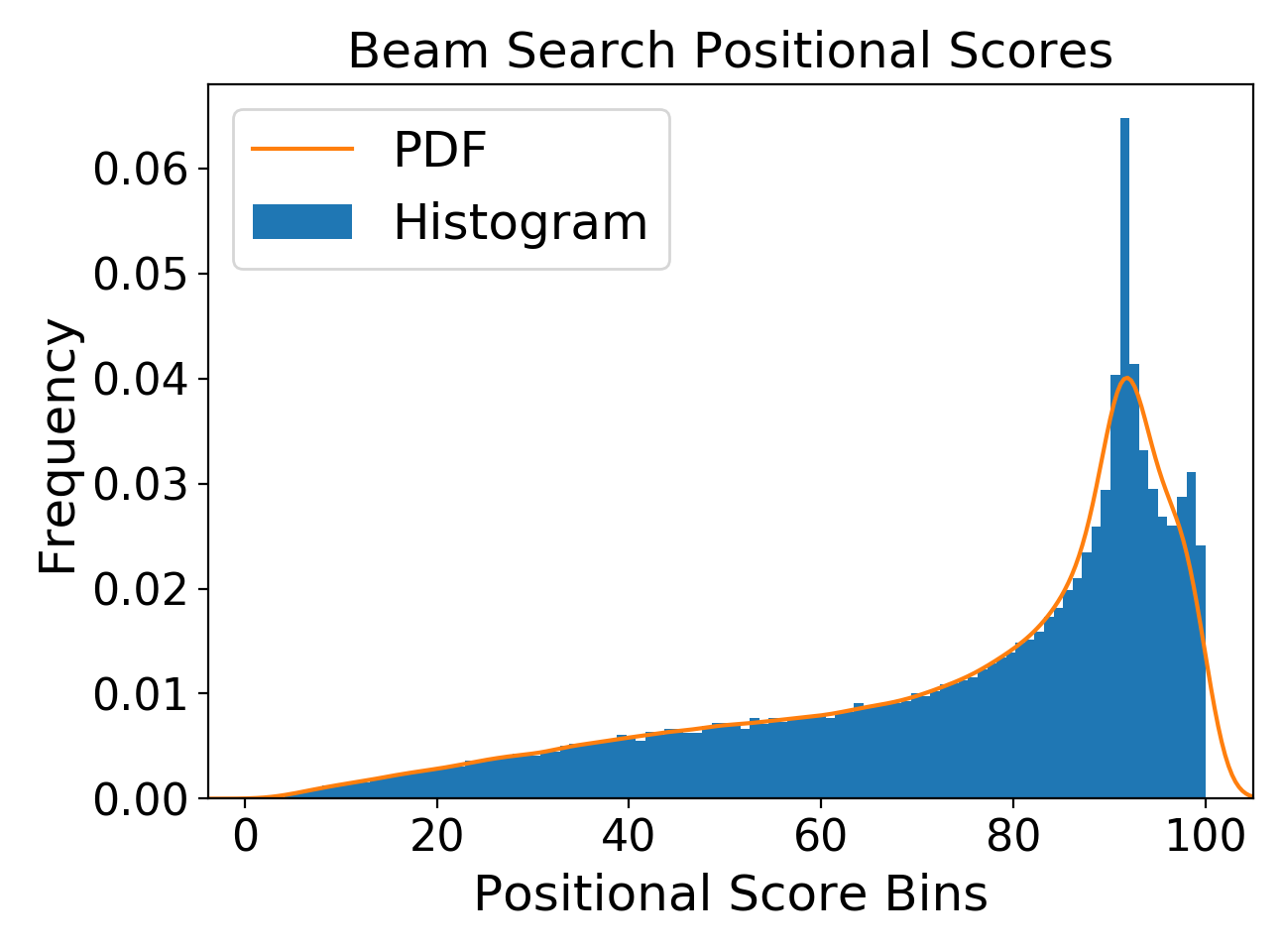}
       \includegraphics[width=0.48\textwidth,height=0.3\textwidth]{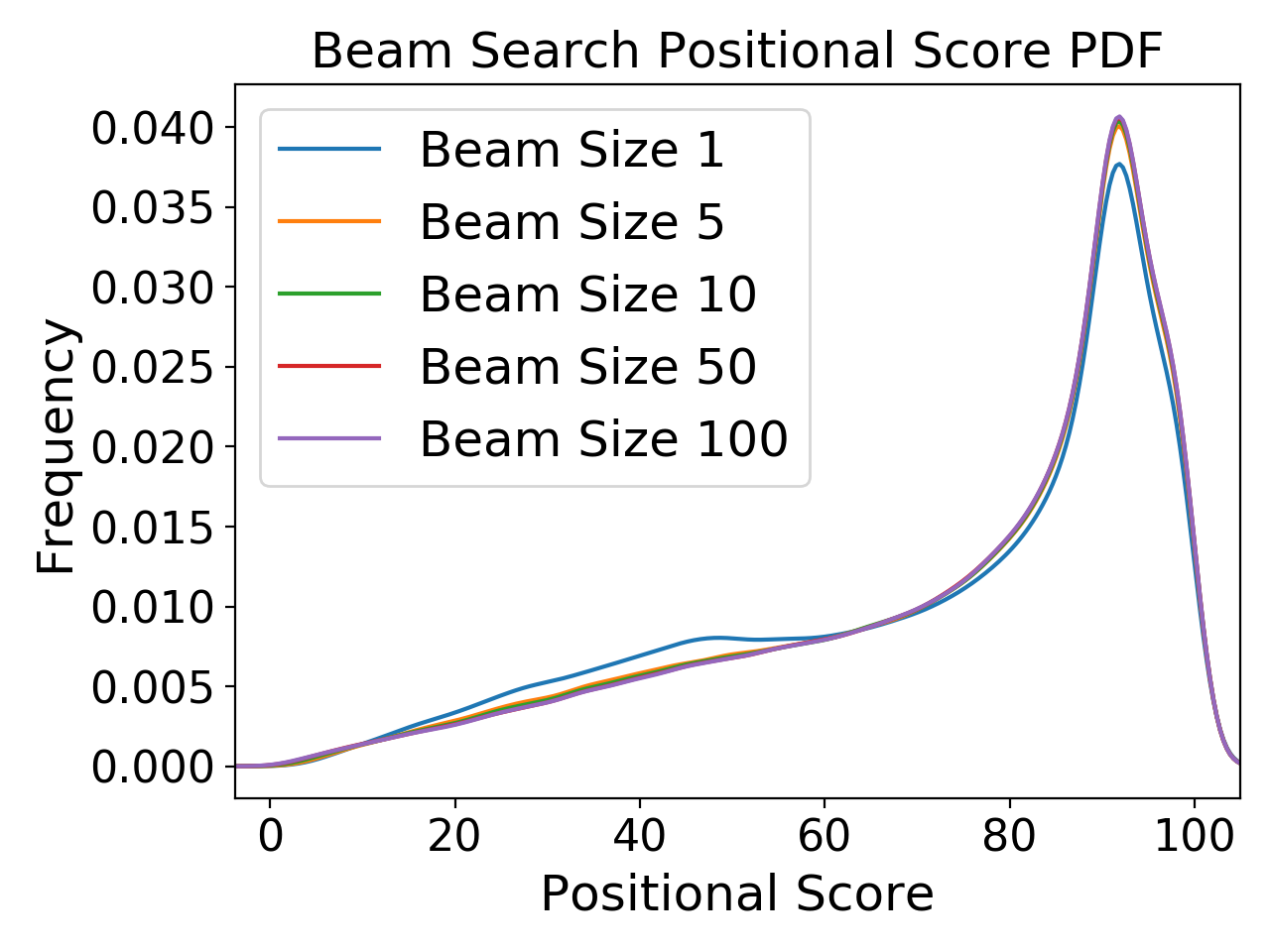}
    \caption{Beam Search Analysis: (Top) Positional scores for Beam size = $5$ and (Bottom) PDFs for different Beam sizes. Scores aggregated over the validation set using the IWSLT 14 De-En Model from section \ref{sec:quantify}.}
    \label{fig:figure2}
    \vspace{-1.5em}
\end{figure}

\section{Modeling the Long Tail}
To improve the generation of the long-tail of low frequency tokens, it is important to study how low-frequency tokens could appear in the candidate hypotheses during search. Subsequently, we could leverage any such \textit{biases} from sequence level inference to better model token classification.

\vspace{0.25em}
\noindent \textbf{Beam Search Analysis} To better establish the link between token level classification and beam search inference, we study the distribution of \textit{positional scores}, i.e. the probabilities selected during each step of decoding, for the top hypothesis finally selected during beam search. The top plot in Figure \ref{fig:figure2} shows the histogram of the positional scores, aggregated on the validation set. A Gaussian Kernel density estimator is fitted to the histograms as well, and probability density functions (PDFs) for positional scores are plotted for different beam sizes in Figure \ref{fig:figure2} (the bottom plot).

An analysis of the positional scores (Figure \ref{fig:figure2}, top) reveals that approximately 40 $\%$ of the tokens selected in the top hypothesis have probabilities below 0.75. Further, the bottom plot in Figure \ref{fig:figure2} shows that this distribution is consistent across different beam sizes. These observations show that \textit{the approximate inference procedure of beam-search relies significantly on low confidence predictions}. However, if low-confidence predictions are excessively penalized, the conditional probability distribution will be pushed to lower and lower entropy, hurting effective search. Therefore, we argue that \textit{a better trade-off between token level classification and sequence level inference in NMT could be established by allowing low-confidence predictions to suffer less penalization vis-à-vis cross-entropy.}

\begin{figure}[ht]
    \vskip -2pt
    \centering
       \includegraphics[width=0.48\textwidth]{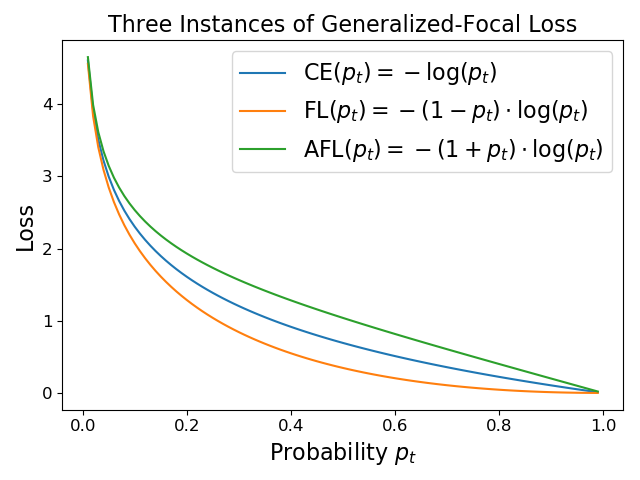}
    \vspace{-1.5em}
    \caption{Comparison of the Loss Functions: Focal loss penalizes low-confidence predictions most aggressively, while Anti-Focal loss relaxes the relative loss for low-confidence predictions vis-à-vis cross-entropy.}
    \label{fig:losses}
    \label{fig:figure3}
    \vspace{-1.0em}
\end{figure}
\begin{figure}[ht!]
    \vskip -2pt
    \centering
    \includegraphics[width=0.47\textwidth,height=0.4\textwidth]{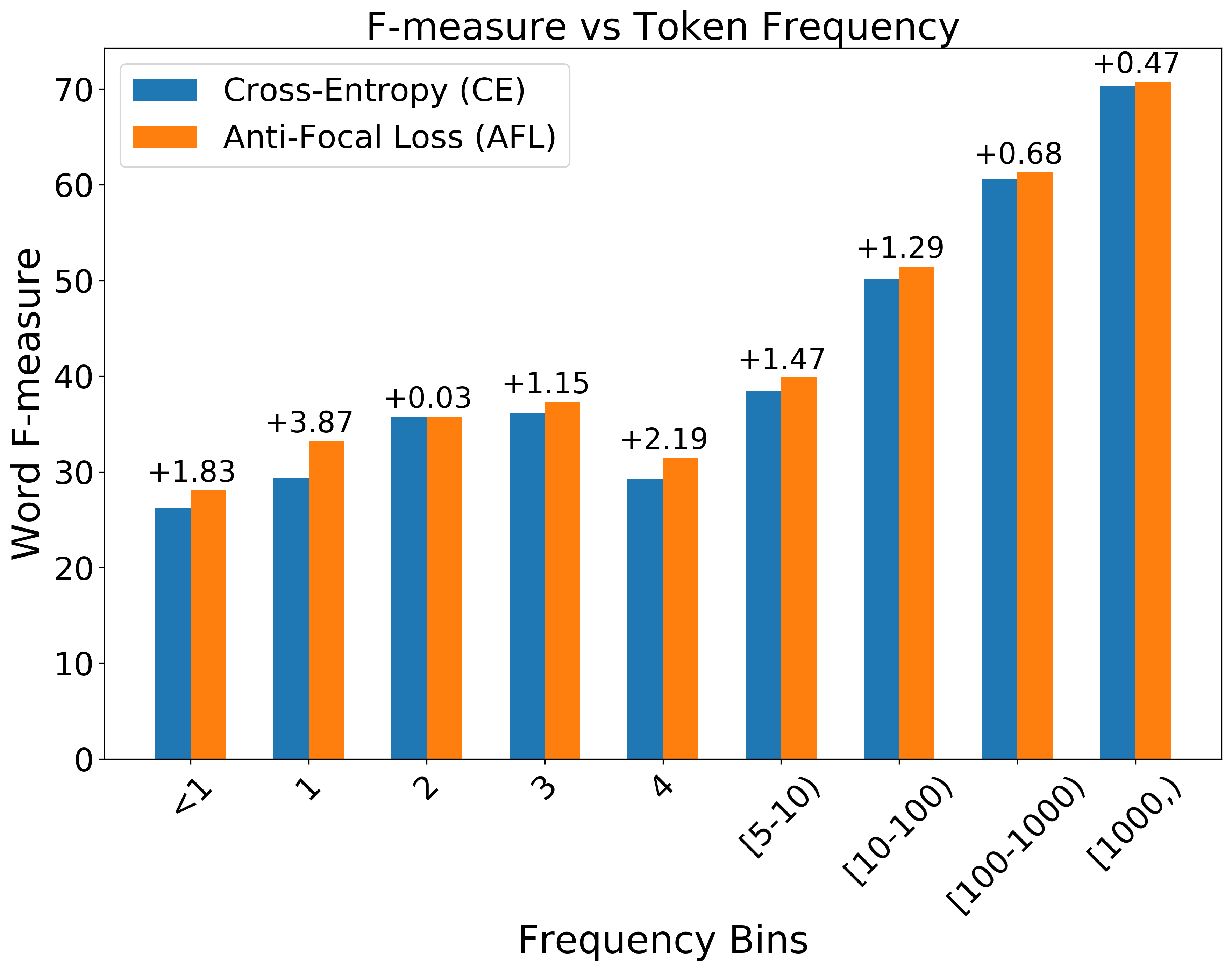} 
    \vspace{-0.75em}
    \caption{Test Word F-measure bucketed by Training Frequency: AFL leads to gains in F-measure across different frequency bins, especially in low-frequency bins.}
    \label{fig:figure4}
    \vspace{-1.0em}
\end{figure}

\begin{table*}[ht]
  \centering
  \setlength\tabcolsep{5.0pt}
\begin{tabular}{ccccccccccc} 
    \toprule
            &    &  & \multicolumn{2}{c} {FL} & \multicolumn{2}{c} {CE + $\tau$-Norm}  &
                \multicolumn{2}{c}{AFL} & \multicolumn{1}{c} {AFL + $\tau$-Norm}  \\
    \cmidrule(r){4-5}\cmidrule(r){6-7}\cmidrule(r){8-9}\cmidrule(r){10-10}
    Dataset & Pair &  CE    & $\gamma=1$          & $\gamma=2$      & $\tau=0.2$   & $\tau=0.4$     & $\alpha=0.5$ & $\alpha=1$ & $\alpha = 1$, $\tau= 0.2$  \\
    \midrule 
    IWSLT 14 & De-En & 32.15   &    31.53    & 30.60 & 32.62 &  32.48  & 32.95 & 33.17 & \textbf{33.41}      \\
    IWSLT 14 & En-De & 26.93  & 26.27  & 25.35 & 27.16 & 26.69  & \textbf{27.35} & 27.05  & 27.31   \\
    IWSLT 14 & Es-En  & 38.95   & 38.30  & 37.41 & 39.29  &  39.28      & 39.33 & 39.47   & \textbf{39.86}    \\
    \midrule
     IWSLT 17 &  En-Fr &  34.40    &  34.60  &  32.60  & 34.30 &  33.70    & \textbf{35.40} &   34.90  & 34.80   \\  
    IWSLT 17 & Fr-En    & 34.60     &  34.00        & 33.30     & 35.10  &      34.60   & 35.00 & 35.00 & \textbf{35.30 }  \\ \midrule 
    TED Talks  & Ru-En & 25.22   & 24.24 & 23.68  & 25.22 & 24.97  & 25.39 & 25.64 & \textbf{25.70} \\
    TED Talks & Pt-En & 34.31   & 32.78  & 31.17  & 34.68  & 34.56  & 34.43 & 35.06 & \textbf{35.31} \\ \midrule
    TED Talks & Gl-En & 13.66   & 13.53  & 13.26  & 13.86 & 13.73 & \textbf{14.82} & 13.73  & 13.84 \\
     TED Talks  & Be-En & \hphantom{0}3.56   & \hphantom{0}4.01  & \hphantom{0}4.28  & \hphantom{0}3.78  & \hphantom{0}3.92  & \hphantom{0}3.97 & \hphantom{0}4.63 & \hphantom{\textbf{0}}\textbf{4.69} \\
    \bottomrule
  \end{tabular}
  \vspace{-0.25em}
  \caption{Test BLEU Scores of the Baselines \& the Proposed Method. Anti-Focal loss consistently leads to significant gains over cross-entropy, with a $p$-value < $0.01$ for each language pair \cite{multeval}. Here CE, FL, and AFL represent cross-entropy, focal, Anti-focal loss respectively. Validation results are presented in Appendix \ref{sec:supplemental3}.}
\vspace{-1.0em}
  \label{tab:table2}
  
\end{table*}
\noindent \paragraph{Anti-Focal Loss} Now, we try to establish a better trade-off for penalizing low-confidence predictions, which could help improve search, while being simple and automatic. Firstly, we generalize Focal loss by introducing a new term $\alpha$ in equation \ref{focal_loss}: 
\begin{equation} \label{anti_focal_loss}
\vspace{-0.25em}
\text{Generalized-FL}(p) = -(1+ \alpha \cdot p)^\gamma \log(p)
\vspace{-0.25em}
\end{equation}
Clearly, for $\alpha=-1$ and $\gamma>0$, Generalized-FL (equation \ref{anti_focal_loss}) reduces to the Focal loss, while for $\alpha=0$, it reduces to the cross-entropy loss. Since we intend to increase the entropy of the conditional token classifier in NMT, we propose to use Generalized-FL with $\alpha>0$ and $\gamma>0$, which we name as \textit{Anti-Focal loss} (AFL). To understand how AFL realizes the intuition derived through beam search analysis, consider Figure \ref{fig:figure3}. Figure \ref{fig:figure3} shows the plot for CE, FL with $\gamma=1$ and AFL with $\gamma=1$ and $\alpha=1$. In general, AFL allocates less relative loss to low-confidence predictions. For example, if we compare the relative loss term  $\cfrac{loss(p=0.6)}{loss(p=0.9)}$ for the three different losses in Figure \ref{fig:figure3}, then CE has a score of 4.85, FL has a score of 19.39, while AFL has a score of 4.08. Further, using $\alpha$ and $\gamma$, we can manipulate the relative loss. Empirically, we find that $\gamma=1$ and $\alpha \in \{0.5, 1.0\}$ works well for AFL in practice.

\section{Experiments and Results}
\label{sec:section5}

We evaluate our proposed Anti-Focal loss against different baselines (CE, FL, $\tau$-Norm) on the task of NMT and analyze the results for further insights.
\paragraph{Datasets and Baselines} We evaluate the proposed algorithm on the widely studied IWSLT 14, IWSLT 17 \cite{cettolo2017overview} and the Multilingual TED Talks datasets \cite{neubig} (details in Appendix \ref{sec:supplemental1}). For model training, we replicate the hyperparameter settings of \citet{incorporating_bert}, except that we do \textit{not} include label-smoothing for a fair comparison of the loss functions (CE, FL, AFL).  $\gamma=1$ is set for AFL. Further, $\tau$-Normalization ($\tau$-Norm) was applied post-training both for CE, AFL. Hyperparameters $\gamma, \alpha, \tau$ were manually tuned.
\paragraph{Experimental Settings} For experiments, we use fairseq~\cite{fairseq}  (more details in Appendix \ref{sec:supplemental2}). For each language pair, BPE with a joint token vocabulary of 10K was applied over tokenized text. A six-layer Transformer model with embedding size 512, FFN layer dimension 1024 and 4 attention heads (42M parameters), was trained for 50K updates for IWSLT datasets and 40K updates for TED Talks datasets. A batch size of 4K tokens, dropout of 0.3 and tied encoder-decoder embeddings were used. BLEU evaluation (tokenized) for IWSLT 14 and TED talks datasets is done using multi-bleu.perl\footnote{\url{https://bit.ly/2Xyst5b}}, while for IWSLT 17 datasets SacreBLEU is used \cite{sacrebleu}. All models were trained on one Nvidia 2080Ti GPU and a beam size of 5 was used for each evaluation.

\paragraph{Results} The trends in Table \ref{tab:table2} show that AFL consistently leads to significant gains over cross-entropy. Further, in Table \ref{tab:mtresult2} we compare CE and AFL ($\alpha=1$) for the three \textit{validation} splits created in section \ref{seqlevel}, for the IWSLT 14 De-En dataset. Table \ref{tab:mtresult2} shows that AFL improves the model the most on the split with the least $\hat{F_{S}}$, while leading to consistent gains on all the three splits.  

Further, Figure \ref{fig:figure4} shows that AFL also leads to gains in word F-measure across different low-frequency bins (evaluated on the test set), implying better generation of low-frequency words. Here, the analysis was done on semantically meaningful word units, using the generated output after the BPE merge operations. Figure \ref{fig:figure_bpe} in Appendix \ref{sec:supplemental4} shows that similar trend holds true for BPE tokens as well. 
Table \ref{tab:table2} also shows that $\tau$-Normalization helps improve BLEU for both CE and AFL, except on En-Fr, providing a simple way to improve NMT models. In general, $\tau$-Norm + AFL leads to the best BLEU scores in Table 2.
\paragraph{Discussion.}  The results show that AFL ameliorates low-frequency word generation in NMT, leading to improvements for long-tailed phenomena both at the token and sentence level. Further, on the two very low-resource language pairs of Be-En and Gl-En, FL leads to improvements, suggesting that under severely poor conditional modeling i.e token classification, explicitly improving long-tailed token classification helps sequence generation in NMT. However, since FL is more aggressive than CE in pushing low-confidence predictions to higher confidence values, in high-resource pairs (with better token classification), FL ends up hurting beam search. Conversely, AFL achieves significant gains in BLEU scores by incorporating the inductive biases of beam search, e.g. in the comparatively higher-resource IWSLT-17 En-Fr dataset (237K training sentence pairs). Here, we also hypothesize that the long-tailed phenomena have considerably different characteristics for low-resource and high-resource language pairs, but leave further analysis for future work.

\begin{table}
  \label{tab:table3}
  \centering
  \small
\setlength\tabcolsep{2.0pt}
  \begin{tabular}{lccccc}
    \toprule
    \textbf{Split} & \textbf{Loss} &\textbf{BLEU $\uparrow$} & \textbf{METEOR $\uparrow$} & \textbf{TER $\downarrow$} & \textbf{R-BERT $\uparrow$} \\
    \midrule
    Highest & CE & 36.7 & 35.5 & 41.7 & 64.0 \\
    Highest & AFL & \textbf{37.1} & \textbf{35.7} & \textbf{41.4} & \textbf{64.3} \\ \midrule
    Medium & CE & 32.3 & 33.3 & 46.3 & 59.9 \\
    Medium & AFL & \textbf{33.3} & \textbf{33.6} & \textbf{45.4} & \textbf{60.5} \\ \midrule
    Least & CE & 31.3 & 33.2 & 46.9 & 59.7 \\
    Least & AFL & \textbf{32.1} & \textbf{33.5} & \textbf{46.4} & \textbf{60.4} \\
    \midrule
  \end{tabular}
  \vspace{-0.75em}
  \caption{Sequence Level Long-Tailed Phenomena: CE vs AFL for different MT metrics, for IWSLT 14 De-En.}
  \label{tab:mtresult2}
  \vspace{-2.0em}
\end{table}

\section{Conclusion and Future Work}
\label{sec:evaluation}

In this work, we characterized the long-tailed phenomena in NMT and demonstrated that NMT models aren't able to effectively generate low-frequency tokens in the output. We proposed a new loss function, the Anti-Focal loss, to incorporate the inductive biases of beam search into the NMT training process. We conducted comprehensive evaluations on 9 language pairs with different amounts of training data from the IWSLT and TED corpora. Our proposed technique leads to gains across a range of metrics, improving long-tailed NMT at both the token as well as at the sequence level. In future, we wish to explore its connections to entropy regularization and model calibration and whether we can fully encode the inductive biases of label smoothing in the loss function itself.

\section*{Acknowledgments}
This research was supported in part by DARPA grant FA8750-18-2-0018 funded under the AIDA program and the DARPA KAIROS program from the Air Force Research Laboratory under agreement number FA8750-19-2-0200. 
The U.S. Government is authorized to reproduce and distribute reprints for Governmental purposes not withstanding any copyright notation there on. The views and conclusions contained herein are those of the authors and should not be interpreted as necessarily representing the official policies or endorsements, either expressed or implied, of the Air Force Research Laboratory or the U.S. Government.

\bibliography{longtail_2284_emnlp2020}
\bibliographystyle{acl_natbib}

\appendix
\section{Dataset Statistics}
\label{sec:supplemental1}
The dataset statistics are highlighted in Table \ref{tab:dataset_statitiscs}, while descriptions of the language pairs are provided in Table \ref{tab:datasetpairsappendix}. The preparation of validation and test sets for IWSLT 14 and 17 datasets is done using fairseq \cite{fairseq} scripts, following \citet{incorporating_bert} \footnote{\url{https://bit.ly/2MtV2tW}} for the corresponding datasets. The TED talks dataset is provided with train, validation and test sets \cite{neubig}. Further, the TED talks dataset is tokenized using moses, and the data preparation script is based on the IWSLT 14 data preparation script in fairseq. We have provided the data preparation scripts as well, from download to pre-processing for each of the datasets, in the code.

\begin{table*}[!htbp]
  \centering
     \setlength\tabcolsep{5.0pt}
\begin{tabular}{ccccccccccc} 
    \toprule
            &    &  & \multicolumn{2}{c} {FL} & \multicolumn{2}{c} {CE + $\tau$-Norm}  &
                \multicolumn{2}{c}{AFL} & \multicolumn{1}{c} {AFL + $\tau$-Norm}  \\
    \cmidrule(r){4-5}\cmidrule(r){6-7}\cmidrule(r){8-9}\cmidrule(r){10-10}
    Dataset & Pair &  CE    & $\gamma=1$          & $\gamma=2$      & $\tau=0.2$   & $\tau=0.4$     & $\alpha=0.5$ & $\alpha=1$ & $\alpha = 1$, $\tau= 0.2$  \\
    \midrule 
    IWSLT 14 & De-En & 33.44   &    33.39    & 33.17 & 33.86 &  33.64  & 33.98 & 34.34 & \textbf{34.64}      \\
    IWSLT 14 & En-De & 28.02  & 27.22  & 26.37 & 27.87 & 27.20  & \textbf{28.35} & 28.23  & 28.27   \\
    IWSLT 14 & Es-En  & 41.19 & 40.36  & 39.31 & 41.39  & 41.08       & 41.26 & 41.26   & \textbf{41.65}    \\
    \midrule
     IWSLT 17 & En-Fr &  34.60  &  33.67  & 33.67  & 33.98 &  33.41    & \textbf{34.81} &   33.93  & 34.12   \\  
    IWSLT 17 & Fr-En    & 32.73     &  32.16        & 31.89     & 33.05  &      32.82   & 32.56 & 33.12 & \textbf{33.16 }  \\ \midrule
    TED Talks & Ru-En & 25.50   & 24.85 & 24.33  & 25.65 & 24.67 & 25.86 & 25.88 & \textbf{25.95}  \\
    TED Talks & Pt-En & 35.34   & 33.42  & 32.47  & 35.47  & 35.09 & 35.68 & 35.71 & \textbf{36.05} \\ \midrule
     TED Talks & Be-En & \hphantom{0}4.24  & \hphantom{0}5.25  & \hphantom{0}5.43  & \hphantom{0}4.54  & \hphantom{0}4.39 & \hphantom{0}5.46 & \hphantom{0}5.35 & \hphantom{\textbf{0}}\textbf{5.71} \\
      TED Talks & Gl-En & 14.64   & 14.66  & 13.54  & 14.95 & 14.87 & \textbf{15.72} & 15.17 & 14.97 \\
    \bottomrule
  \end{tabular}
  \caption{BLEU Scores of the Baselines and the Proposed Method on the \textit{Validation} set.}
  \label{tab:table2appendix}
\end{table*}

\begin{table}[ht!]
  \centering
   \setlength\tabcolsep{4.5pt}
  \begin{tabular}{cccc} 
    \toprule
    Dataset & Source & Target  & Lang-Pair  \\
    \midrule 
    IWSLT 14 & German & English   &  De-En       \\
    IWSLT 14 & English & German   & En-De \\
    IWSLT 14 & Spanish & English & Es-En \\
    \midrule
    IWSLT 17 & English & French    &  En-Fr   \\  
    IWSLT 17 & French & English & Fr-En \\ \midrule
    TED Talks & Russian & English  & Ru-En\\
    TED Talks & Portuguese & English  & Pt-En \\ \midrule
     TED Talks & Belarusian & English  & Be-En \\
      TED Talks & Galician & English  & Gl-En \\
    \bottomrule
  \end{tabular}
  \label{tab:datasetpairdetails}
\caption{Dataset Language Pair Details: The abbreviations for the language pairs are used throughout.}
\label{tab:datasetpairsappendix}
\end{table}

\begin{table}[ht!]
    \centering
    \setlength\tabcolsep{4.5pt}
    \begin{tabular}{ccccc}
    \toprule 
    Dataset & Pairs & Train & Valid & Test\\
    \midrule
    IWSLT 14 & En-De& 160,239 & 7,283 & 6,750\\
    IWSLT 14 & De-En & 160,239 & 7,283 & 6,750\\
    IWSLT 14 & Es-En & 169,028 & 7,683 & 5,593\\
    \midrule
    IWSLT 17 & En-Fr & 236,652 & 890 & 1,210 \\
    IWSLT 17 & Fr-En & 236,652 & 890 & 1,210 \\
    \midrule
    TED Talks & Ru-En & 208,106 & 4,805 & 5,476 \\
    TED Talks  & Pt-En & \hphantom{0}51,785 & 1,193 & 1,803 \\
    TED Talks  & Gl-En & \hphantom{0}10,017 & \hphantom{0}682 & 1,007 \\
    TED Talks & Be-En & \hphantom{0}\hphantom{0}4,509 & \hphantom{0}248 & \hphantom{0}664 \\
    
     \bottomrule
    \end{tabular}
    \caption{Dataset Statistics: Train, Validation and Test Splits for each of the Language Pairs.}
    \label{tab:dataset_statitiscs}
\end{table}

\section{Model Details}
\label{sec:supplemental2}

The Transformer model is the \texttt{iwslt-de-en} model architecture in fairseq \footnote{\url{https://bit.ly/3dxfOoB}}, also used in \citet{incorporating_bert}. It is a six-layer Transformer model (6 layers in both the encoder and decoder) with embedding size 512, FFN layer dimension 1024 and 4 attention heads. The optimizer used is Adam, with a learning rate of 0.0005, with 4K warmup updates a warmup initial learning rate of $1e-07$. We have provided training as well as evaluation scripts for each of the datasets in the code. The loss functions are implemented by subclassing cross-entropy in the fairseq framework and are available in the \texttt{Criterions} directory.

\section{Validation Results}
\label{sec:supplemental3}
Table \ref{tab:table2appendix} provides the results for the \textit{Validation} set, corresponding to the test set evaluation done in Table 2 in section 5 of the main paper. The evaluation settings remain the same as in Section 5, except that, the validation results for IWSLT 17 are obtained using multi-bleu.perl\footnote{\url{https://bit.ly/2Xyst5b}} instead of SacreBLEU \cite{sacrebleu}. In general, \textit{Validation} set results also adhere to the same trend as in Section 5. In particular, Anti-Focal, combined with $\tau$-Normalization (AFC + $\tau$-Norm) leads to gains in cross-entropy over each of the datasets.

\section{F-Measure Comparison}
\label{sec:supplemental4}
\begin{figure}[ht!]
    \vskip -2pt
    \centering
    \includegraphics[width=0.48\textwidth,height=0.43\textwidth]{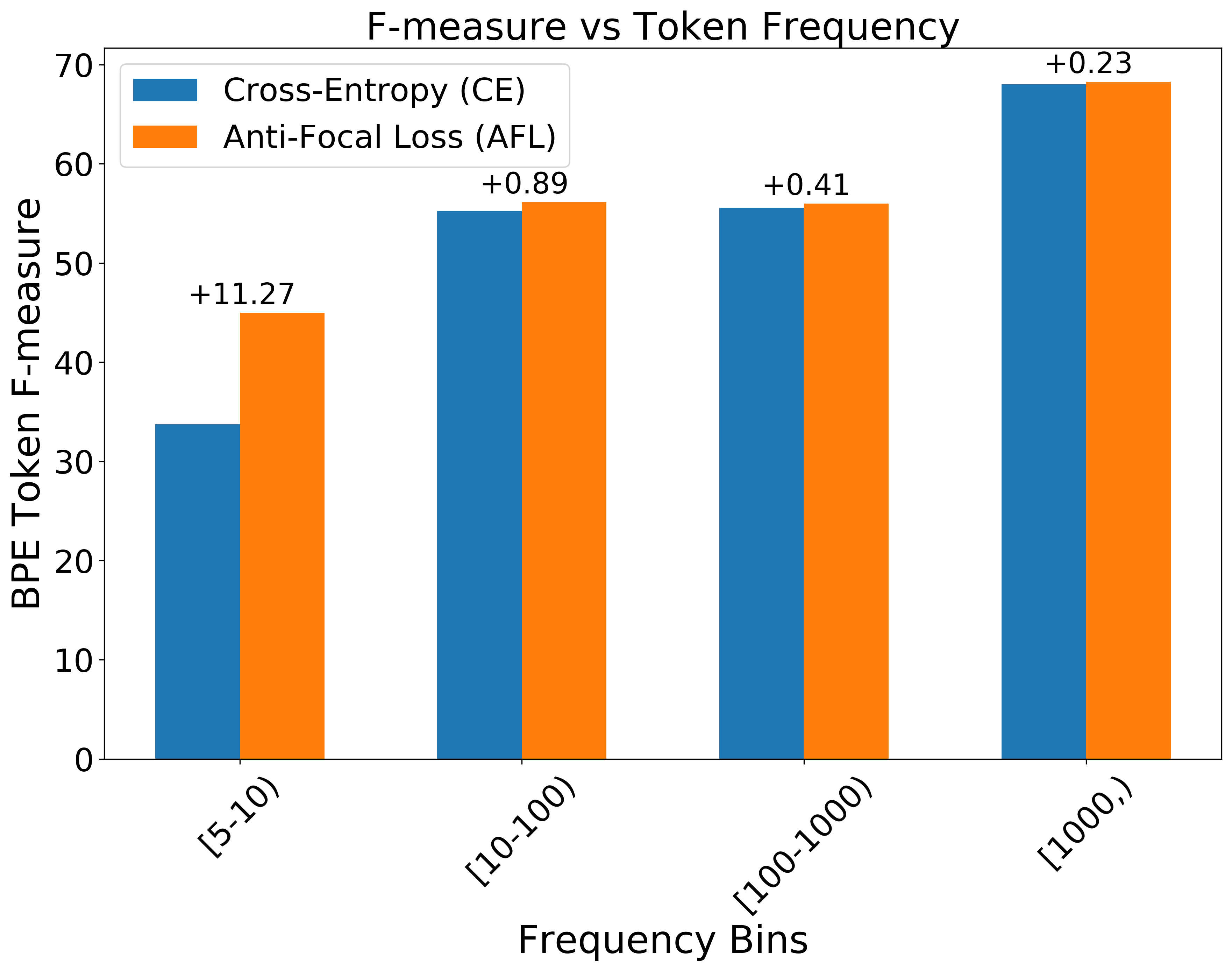} 
    \vspace{-0.75em}
    \caption{Test F-measure for BPE tokens bucketed by Training Frequency: AFL leads to gains in F-measure across different frequency bins, especially in low-frequency bins.}
    \label{fig:figure_bpe}
    \vspace{-1.0em}
\end{figure}

Figure \ref{fig:figure_bpe} presents the token-level comparison on the generated output without merging the BPE tokens, i.e. Figure \ref{fig:figure_bpe} is the BPE token analogue of Figure \ref{fig:figure4} in Section \ref{sec:section5}. Here also, we observe similar trend for AFL, i.e. AFL leads to considerable gains in F-measure in the lower frequency buckets (e.g. [5-10)), when compared to cross-entropy.

\end{document}